\documentclass[10pt,twocolumn]{article}

\usepackage[utf8]{inputenc}
\usepackage[T1]{fontenc}
\usepackage{times}
\usepackage[margin=0.75in]{geometry}
\usepackage{microtype}
\usepackage{graphicx}
\usepackage{booktabs}
\usepackage{multirow}
\usepackage{array}
\usepackage{tabularx}
\usepackage{amsmath,amssymb,amsfonts}
\usepackage{mathtools}
\usepackage[numbers,sort&compress]{natbib}
\usepackage{xcolor}
\usepackage[hidelinks]{hyperref}
\usepackage{caption}
\usepackage{subcaption}
\usepackage{enumitem}
\usepackage{url}
\usepackage{placeins}
\usepackage{float}

\graphicspath{{figures/}}

\setlist[itemize]{leftmargin=*,topsep=2pt,itemsep=2pt,parsep=0pt}

\newcommand{\R}{\mathbb{R}}
\newcommand{\C}{\mathbb{C}}
\newcommand{\Sphere}{\mathbb{S}^{2}}

\newcommand{\PSL}{\mathrm{PSL}(2,\mathbb{C})}
\newcommand{\Kqc}{K}

\newcommand{\degree}{\ensuremath{^\circ}}

\newcommand{\na}{\textcolor{gray}{n/a}}

\title{\vspace{-0.5em}\textbf{Shear-Free Viewport Magnification for 360\degree{} Images\\[-0.15em]
via Spherical M\"obius Boosts}}

\author{
Boyang Li \qquad Hezhao Xu\\
Peking University
}
\date{}

\begin{document}
\maketitle

\begin{abstract}
Viewport-adaptive 360\degree{} imaging seeks to allocate a fixed sampling budget to the region a viewer is likely to observe. Existing view-biased projections increase viewport resolution through non-conformal warps, which can introduce anisotropic stretching and shear. We formulate spherical M\"obius boosts as exact conformal maps for fixed-budget viewport magnification. The continuous spherical warp has quasiconformal dilatation $\Kqc=1$, reallocating samples toward a target direction while preserving local angles. On a SUN360 saliency audit with 72 panoramas and 216 paired viewport targets, C1 M\"obius boosting improves viewport PSNR over optimized offset cubemap on all paired cases, with case-level median gain $+3.26$ dB, image-level median gain $+3.23$ dB, and panorama-level bootstrap 95\% CI $[+3.15,+3.33]$ dB. Pareto analysis shows that this is not a free global-quality improvement: C1 trades full-sphere WS-PSNR for shear-free viewport fidelity. Prediction-error and filtering studies identify the operating envelope: strong boosts are useful for accurately targeted viewports, while large target uncertainty calls for weaker boosts or fallback. These results position M\"obius boosting as a geometric primitive for prediction-conditioned foveated 360\degree{} resampling rather than a universal encode-once layout.
\end{abstract}

\begin{figure*}[t]
    \centering
    \includegraphics[width=\textwidth]{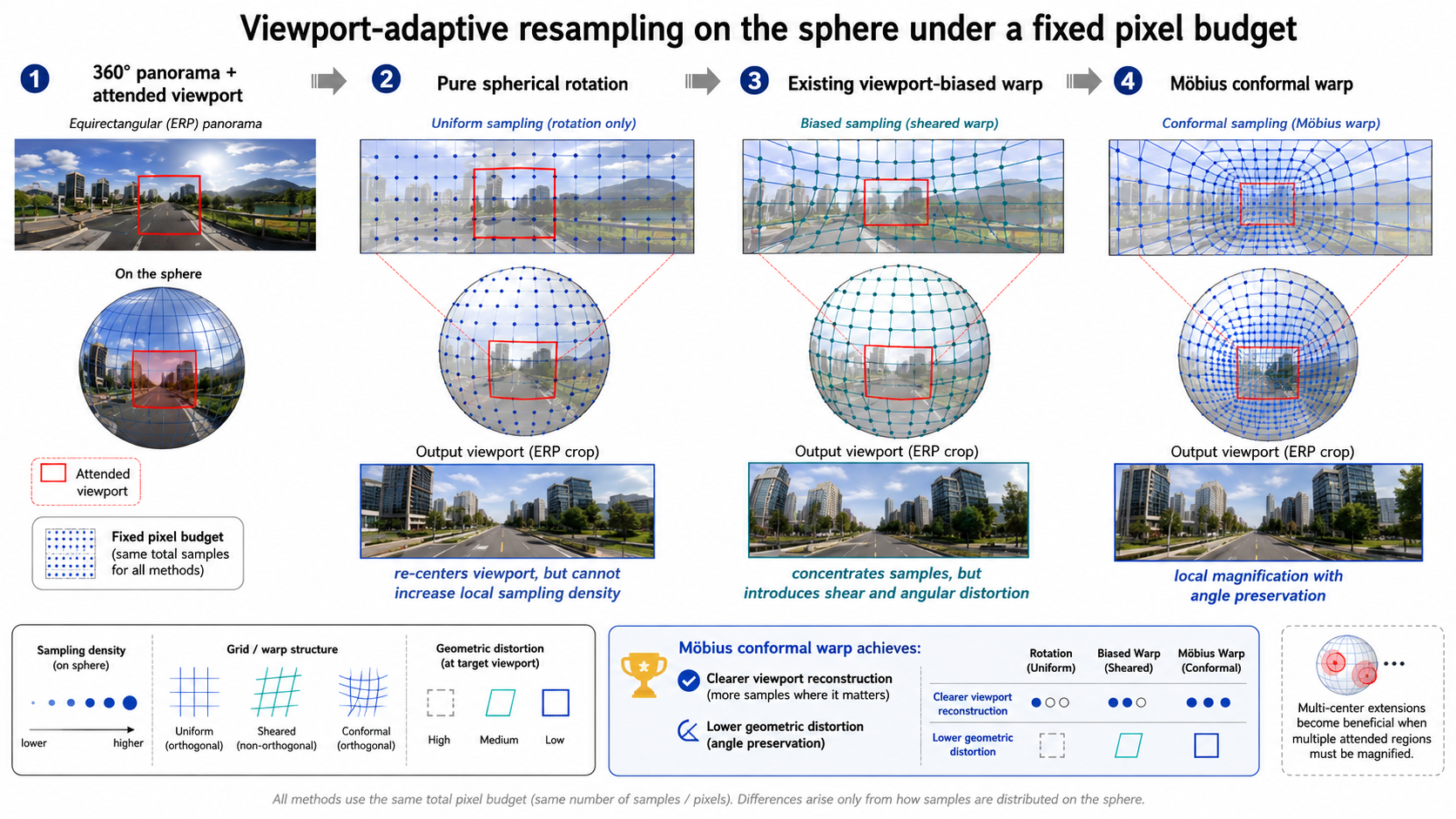}
    \caption{\textbf{Viewport-adaptive resampling on the sphere under a fixed pixel budget.}
    A pure spherical rotation can re-center the attended viewport but cannot increase its local sampling density. Existing viewport-biased warps can concentrate samples, but may introduce shear and angular distortion. A M\"obius conformal warp reallocates samples toward the viewport while preserving local angles, giving a clean primitive for foveated 360\degree{} resampling.}
    \label{fig:teaser}
\end{figure*}

\section{Introduction}
A 360\degree{} image gives the viewer freedom to look in any direction, but a fixed raster budget cannot represent all directions with equal perceptual value. Most viewers occupy a small instantaneous viewport, and immersive systems exploit this by predicting head or gaze direction and allocating more bitrate or samples near the likely view. The geometric question behind such systems is simple: how should a finite set of samples on the sphere be redistributed when the target viewport is known or predicted?

Standard spherical layouts only partially answer this question. Equirectangular projection (ERP) is simple but oversamples the poles; cube maps reduce polar waste but create face seams; tangent images reduce local projection distortion by using many locally planar charts~\citep{eder2020tangent}. These layouts are mostly content- and viewer-agnostic. They change the baseline parameterization of the sphere, but they do not deliberately move samples toward one viewer's current viewport.

Viewport-biased projections address the mismatch more directly, but their geometric mechanism matters. A pure rotation can place content in a favorable orientation, yet it is an isometry of the sphere and therefore cannot increase local sampling density. Offset cubemap, pyramid, truncated-square-pyramid, and related view-dependent layouts can increase viewport density, but they do so through non-conformal warps that may introduce anisotropic stretching and shear. Figure~\ref{fig:teaser} illustrates the resulting gap: rotation cannot magnify, while generic viewport magnification can damage local shape.

Our key observation is classical. The orientation-preserving conformal automorphism group of the sphere is the M\"obius group $\PSL$. It strictly contains the rotation group. In addition to rotations, it includes boost-like degrees of freedom that magnify one spherical neighborhood while compressing the opposite side. Since a M\"obius map is conformal, its local tangential Jacobian is a scalar times a rotation. Thus it changes local scale without changing local angles. This makes the M\"obius boost a natural fixed-budget primitive for shear-free viewport magnification.

This paper is deliberately scoped. We do not claim that M\"obius transformations are new to panoramic imaging: they have been used for panoramic display correction, movement and zoom on omnidirectional images, and panoramic-depth augmentation~\citep{penaranda2015mobius,cao2023omnizoomer,cao2025panda}. Our claim is narrower: when the objective is fixed-budget viewport-adaptive resampling, exact single-center M\"obius boosts provide a geometry-matched conformal alternative to non-conformal viewport-biased projections. The expanded experiments make the tradeoff explicit: M\"obius boosts are a strong viewport-fidelity and shear-free operating point, not a free improvement in full-sphere quality.

We make three contributions:
\begin{itemize}
    \item \textbf{Exact conformal viewport magnification.} We derive spherical M\"obius boosts as fixed-budget viewport magnification maps and show that the continuous spherical warp has quasiconformal dilatation $\Kqc=1$, eliminating warp-induced anisotropic distortion.
    \item \textbf{Pareto-aware evidence against non-conformal projection.} Under matched buffers, target directions, and scalar search budgets, we compare C1 against ERP, rotation, and offset cubemap, reporting both viewport fidelity and full-sphere WS-PSNR cost.
    \item \textbf{Operating envelope for foveated deployment.} Across SUN360 saliency targets, prediction-error sweeps, filtering controls, and a small real-gaze trace study, we show that C1 is most useful for concentrated and accurately targeted viewports, while diffuse or uncertain attention requires weakening or fallback.
\end{itemize}

A non-conformal radial focus warp is included as an additional stress-test baseline, but offset cubemap is the primary literature-grounded non-conformal viewport-biased baseline. A windowed multi-center quasiconformal flow is reported only as an appendix extension for far-apart simultaneous attention clusters.

\section{Related Work}
\paragraph{Spherical projections and 360\degree{} representations.}
ERP, cube maps, equal-area variants, and tangent images are standard ways to rasterize spherical images. Tangent images render the spherical signal to locally planar grids tangent to an icosahedral subdivision and reduce spherical distortion for computer-vision pipelines~\citep{eder2020tangent}. VAR studies viewport-adaptive spherical image resampling and interpolation quality~\citep{regensky2023var}. These works motivate careful spherical resampling, but they do not focus on exact conformal viewport magnification under a fixed pixel budget.

\paragraph{Viewport-adaptive projection and streaming.}
Viewport-driven rate-distortion optimization allocates quality based on predicted navigation and content-dependent rate-distortion curves~\citep{chakareski2018viewport}. View-dependent projection families, including pyramid projection, offset cubemap, truncated-square-pyramid projection, and user-adaptive area-of-focus projections, exploit the fact that only part of the sphere is visible at a given time~\citep{kuzyakov2016facebook,zhou2018offset,corbillon2017optimal,elganainy2017tiled,zhou2020adap360}. Tile-based systems instead encode different spatial tiles at different qualities or probabilities~\citep{xie2017prob}. Our work is complementary: it isolates the geometry of the resampling map and asks whether the viewport-biased warp can be made conformal.

\paragraph{Rotation-based methods.}
Snap Angle Prediction predicts rotations that place important content in favorable cube-map locations, reducing seam and distortion artifacts~\citep{xiong2018snap}. Learning Compressible 360\degree{} Video Isomers predicts sphere rotations that improve compressibility under standard encoders~\citep{su2018isomers}. These methods optimize orientation. Since rotations are spherical isometries, they cannot by themselves concentrate samples into a viewport. We use rotations as important baselines but evaluate a different geometric degree of freedom: conformal boost magnification.

\paragraph{M\"obius transformations in panoramic imaging.}
M\"obius transformations have appeared in panoramic display correction and high-resolution omnidirectional movement/zoom~\citep{penaranda2015mobius,cao2023omnizoomer}. PanDA uses M\"obius spatial augmentation to improve panoramic depth estimation from unlabeled panoramas~\citep{cao2025panda}. These works show that the transformation family is useful on spherical imagery. Our contribution is not the existence of M\"obius maps, but their use as a fixed-budget, viewport-conditioned, shear-free resampling primitive and their comparison against non-conformal viewport-biased projection baselines.

\paragraph{360\degree{} quality, gaze, and evaluation.}
WS-PSNR and related spherical metrics account for nonuniform sampling density over the sphere~\citep{sun2017wspsnr}. Viewport-aware quality is essential for immersive media because visible content dominates perceived quality. Salient360 and its toolbox provide head/eye data and processing tools for visual attention on 360\degree{} content~\citep{rai2017salient360,gutierrez2018salient360,david2023toolbox}. We use viewport PSNR for the attended region and WS-PSNR for the whole sphere, because the method intentionally trades global quality for viewport quality.

\section{Method}
\subsection{Fixed-budget resampling on the sphere}
Let $I:\Sphere\rightarrow[0,1]^3$ be a spherical image and let $\mathcal{D}_b$ be a fixed-size buffer grid with the same number of stored samples for all methods. A spherical resampling map $g:\Sphere\rightarrow\Sphere$ defines the encoded buffer by
\begin{equation}
    B[d] = I\big(g^{-1}(d)\big),\quad d\in\mathcal{D}_b,
\end{equation}
and the decoded reconstruction on a high-resolution grid $\mathcal{D}_h$ by
\begin{equation}
    \widehat I[d] = B\big(g(d)\big),\quad d\in\mathcal{D}_h.
\end{equation}
In the continuous limit this is an invertible change of variables. At finite resolution, quality depends on how $g$ distributes samples on the sphere. In implementation, the layout module maps buffer pixels to sphere directions; samples are interpolated on the source raster and on cubemap faces. Face-boundary handling, interpolation kernel, and prefiltering options are fixed within each matched comparison.

\paragraph{Viewport quality.}
For an attention or viewport weight $w(d)\ge0$, we use a solid-angle-weighted RGB MSE,
\begin{equation}
\mathrm{vMSE}(I,\widehat I;w)=
\frac{\sum_{d} w(d)\,\Delta\Omega(d)\,\|I(d)-\widehat I(d)\|_2^2/3}{\sum_{d}w(d)\Delta\Omega(d)},
\end{equation}
where $\Delta\Omega(d)$ is the spherical area element. For ERP samples it is proportional to $\cos\phi(d)$. With RGB values in $[0,1]$, viewport PSNR is
\begin{equation}
    \mathrm{vPSNR}=10\log_{10}\frac{1}{\mathrm{vMSE}}.
\end{equation}
We report WS-PSNR by setting $w\equiv1$ and using the same spherical area weights. This pair of metrics is important: viewport resampling should improve the intended view, but it necessarily spends fewer samples elsewhere.

\subsection{C1: exact M\"obius viewport boost}
We represent an orientation-preserving M\"obius transformation by a matrix $M\in\PSL$ acting on stereographic coordinate $z\in\C\cup\{\infty\}$,
\begin{equation}
    z' = \frac{M_{11}z+M_{12}}{M_{21}z+M_{22}}.
\end{equation}
The Lie algebra parameterization separates rotations and boosts,
\begin{equation}
M=\exp\!\left(\frac12\sum_{k=1}^{3}\left[i r_k\sigma_k+b_k\sigma_k\right]\right),
\end{equation}
where $r\in\R^3$ are rotational parameters, $b\in\R^3$ are boost parameters, and $\sigma_k$ are the Pauli matrices. The rotation-only subgroup is recovered when $b=0$.

For the canonical boost $z'=a z$ with $a>0$, the spherical conformal scale at stereographic radius $\rho=|z|$ is
\begin{equation}
    \lambda_a(\rho)=a\,\frac{1+\rho^2}{1+a^2\rho^2}.
    \label{eq:scale}
\end{equation}
Composing with rotations aims this scale profile at an arbitrary viewport center $c\in\Sphere$. The encoder chooses the sign and magnitude of the boost so that the attended source neighborhood receives higher sample density in the finite buffer. Since every M\"obius map is conformal, the tangential Jacobian is a scaled rotation and
\begin{equation}
    \Kqc(d)=\frac{\sigma_{\max}(J_g(d))}{\sigma_{\min}(J_g(d))}=1\quad\forall d\in\Sphere.
\end{equation}
Thus the continuous C1 spherical warp has zero warp-induced anisotropic distortion: it changes local scale and area, but not local angles. The discrete rasterization pipeline may still introduce interpolation, face-layout, and finite-sampling artifacts; we control these separately in the matched experiments.

\subsection{Matched baselines, Pareto controls, and prediction gating}
The comparison must not reward a method for using more parameters or a larger buffer. Unless stated otherwise, each method receives the same total number of stored samples and the same target direction. The aimed-rotation baseline rotates the target to a canonical view direction and has no magnification parameter. Offset cubemap uses the same six-face buffer and optimizes one scalar offset magnitude in the target direction. We additionally include a non-conformal radial focus warp with a corresponding one-dimensional focus parameter as a stress-test baseline, but we do not treat it as a literature-exact TSP implementation. C1 optimizes one scalar boost strength. All adaptive one-parameter methods use the same grid size and the same oracle objective, viewport PSNR on the shared target viewport; the oracle target is therefore not an advantage for any single method.

We report four matching conditions. The default compares the best viewport PSNR under the same scalar-search budget. A second condition matches center forward scale, isolating whether conformality still helps at comparable local magnification. A third condition matches WS-PSNR cost, asking whether the method improves viewport fidelity at the same full-sphere sacrifice. A fourth condition asks how much additional WS-PSNR cost C1 needs to reach the optimized offset viewport-PSNR target.

For predicted targets, strong viewport magnification is unsafe under large angular error. Our experiments therefore include an error-conditioned gating study: use the strong C1 setting inside a reliable range, weaken the boost as target uncertainty increases, and fall back when uncertainty is too high. This study is not a full viewport predictor; a deployable system would need a predictor confidence estimate to approximate the same rule at test time.

\paragraph{Multi-center extension.}
The main method is the exact one-center C1 boost. A windowed multi-center quasiconformal flow is defined and evaluated only in Appendix~\ref{app:c2}; it is not used as the central evidence for the C1 conformality claim.

\section{Experiments}
The evaluation follows the evidence chain required by the geometric claim: main viewport quality, Pareto tradeoff, angular/shear distortion, robustness to prediction uncertainty, anti-aliasing controls, and a small real-gaze illustrative trace study. Unless noted otherwise, values are mean $\pm$ standard error and $n$ is shown explicitly. Downstream probes, cached runtime, resolution/FoV sanity checks, and the C2 multi-center extension are reported in the appendix because they are not the central evidence for the C1 claim.

\subsection{Datasets and common setup}
Table~\ref{tab:dataset} summarizes the evaluation. The largest still-image audit uses 72 SUN360 test panoramas resized to height $H=256$, with a fixed $120\times60$ buffer for all methods and three saliency-derived viewport centers per image, giving 216 paired cases. A smaller $H=512$ resolution/FoV sanity check is reported in Appendix~\ref{app:sensitivity}.

\begin{table}[t]
\centering
\caption{Dataset coverage in the expanded evaluation. The main SUN360 audit uses $H=256$ panoramas and a fixed $120\times60$ buffer for all compared maps.}
\label{tab:dataset}
\small
\setlength{\tabcolsep}{3.5pt}
\resizebox{\linewidth}{!}{%
\begin{tabular}{lcccc}
\toprule
Dataset / split & Images & Users & View windows & Notes \\
\midrule
Synthetic controlled & generated & 0 & script-defined & Single/multi ROI \\
Real panorama set & 5 & 0 & spectral peak viewports & Static images \\
SUN360 saliency & 72 & 0 & 216 & Aggregate EM saliency, 3 centers/image \\
Salient360 trace study & 1 & 3 & $4\!\times\!5\!\times$users + agg. & Illustrative head/gaze windows \\
\bottomrule
\end{tabular}}
\end{table}

\paragraph{Saliency targets and concentration.}
For SUN360, aggregate EM saliency maps are used only to define static target directions, not as a method-specific quality signal. Each saliency map is resized to the image grid, normalized, and smoothed with a Gaussian kernel ($\sigma=2$ pixels). We select the top three local maxima greedily with spherical non-maximum suppression: a candidate is accepted only if it is at least $35\degree$ from all previously selected centers. Each selected direction defines the same viewport weight for every method, a raised-cosine spherical cap with support radius $25\degree$. Saliency concentration is reported only as an operating-regime descriptor: it is the solid-angle-weighted fraction of aggregate saliency mass inside a hard $25\degree$ spherical cap around the selected center. The same targets, weights, and concentration values are reused for ERP, rotation, offset cubemap, the radial focus warp, and C1.

\paragraph{Scalar optimization protocol.}
All one-parameter adaptive methods optimize viewport PSNR over eight scalar candidates using the same target direction, viewport weight, buffer size, and scalar-search budget. Table~\ref{tab:searchgrid} lists the candidate grids. The selected scalar is used only for the corresponding reconstruction; no method receives additional information beyond the shared target direction and shared evaluation objective.

\begin{table}[H]
\centering
\caption{Scalar search grids used in the expanded SUN360 audit.}
\label{tab:searchgrid}
\scriptsize
\setlength{\tabcolsep}{3pt}
\begin{tabularx}{\linewidth}{@{}lX@{}}
\toprule
Method & Parameter grid \\
\midrule
C1 & \begin{tabular}[t]{@{}l@{}}$s\in\{-3.0,-2.6,-2.2,-1.8,$\\$-1.4,-1.0,-0.6,0.0\}$\end{tabular} \\
Offset & \begin{tabular}[t]{@{}l@{}}$o\in\{-0.95,-0.80,-0.65,-0.50,$\\$-0.35,-0.20,-0.05,0.20\}$\end{tabular} \\
Radial focus & \begin{tabular}[t]{@{}l@{}}$r\in\{0.30,0.45,0.65,0.85,$\\$1.00,1.40,2.10,3.00\}$\end{tabular} \\
\bottomrule
\end{tabularx}
\end{table}

\subsection{Expanded main result}
Table~\ref{tab:mainexpanded} reports the expanded SUN360 audit. C1 is compared against ERP, aimed rotation, and offset cubemap under a fixed buffer and the same scalar-search budget. We also include a non-conformal radial focus warp as a stress-test baseline; it is not used as the literature-grounded opponent for the main claim.

\begin{table*}[t]
\centering
\caption{Expanded geometry-matched comparison on 72 SUN360 panoramas with three saliency centers per image ($n=216$ cases). Values are mean $\pm$ standard error.}
\label{tab:mainexpanded}
\small
\setlength{\tabcolsep}{4pt}
\resizebox{\textwidth}{!}{%
\begin{tabular}{lcccccc}
\toprule
Method & Search dof & vPSNR $\uparrow$ & Gain vs ERP $\uparrow$ & WS-PSNR $\uparrow$ & Paired note & $n$ \\
\midrule
ERP & 0 & $26.23{\pm}0.29$ & $0.00{\pm}0.00$ & $25.80{\pm}0.10$ & reference & 216 \\
Aimed rotation & align & $26.07{\pm}0.29$ & $-0.16{\pm}0.05$ & $26.26{\pm}0.10$ & no magnification & 216 \\
Offset cubemap & 1 & $35.47{\pm}0.28$ & $9.24{\pm}0.09$ & $24.26{\pm}0.11$ & strongest non-conf. baseline & 216 \\
Radial focus warp & 1 & $35.43{\pm}0.31$ & $9.20{\pm}0.11$ & $22.72{\pm}0.11$ & stress-test baseline & 216 \\
\textbf{C1 M\"obius boost} & 1 & $\mathbf{38.71{\pm}0.28}$ & $\mathbf{12.48{\pm}0.10}$ & $19.38{\pm}0.10$ & shear-free viewport point & 216 \\
\bottomrule
\end{tabular}}
\end{table*}

\begin{table}[t]
\centering
\caption{Paired C1 gains on the expanded SUN360 audit. Case wins count all 216 viewport targets; image wins aggregate the three centers within each of the 72 panoramas to avoid pseudo-replication. WS delta is C1 minus Offset.}
\label{tab:paired}
\small
\setlength{\tabcolsep}{3.2pt}
\resizebox{\linewidth}{!}{%
\begin{tabular}{lcccccc}
\toprule
Comparison & Case wins & Image wins & Case med. & Image med. & Image 95\% CI & WS delta \\
\midrule
C1 - Offset & 216/216 & 72/72 & $+3.26$ & $+3.23$ & $[3.15,3.33]$ & $-4.88$ \\
\bottomrule
\end{tabular}}
\vspace{0.25em}
\footnotesize{Mean C1--Offset gain is $+3.24{\pm}0.05$ dB; min/max image-level mean gains are $+2.32/+4.33$ dB.}
\end{table}

\begin{figure*}[t]
\centering
\begin{subfigure}[t]{0.32\textwidth}
\centering
\includegraphics[width=\linewidth]{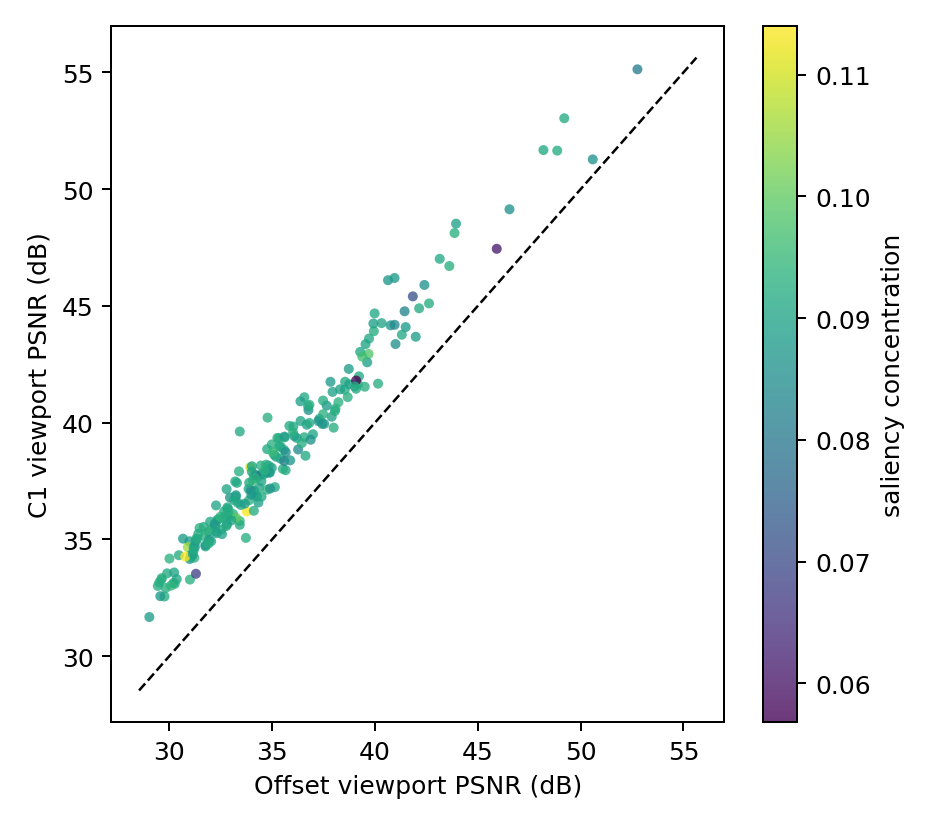}
\caption{C1 vs offset paired vPSNR.}
\end{subfigure}\hfill
\begin{subfigure}[t]{0.32\textwidth}
\centering
\includegraphics[width=\linewidth]{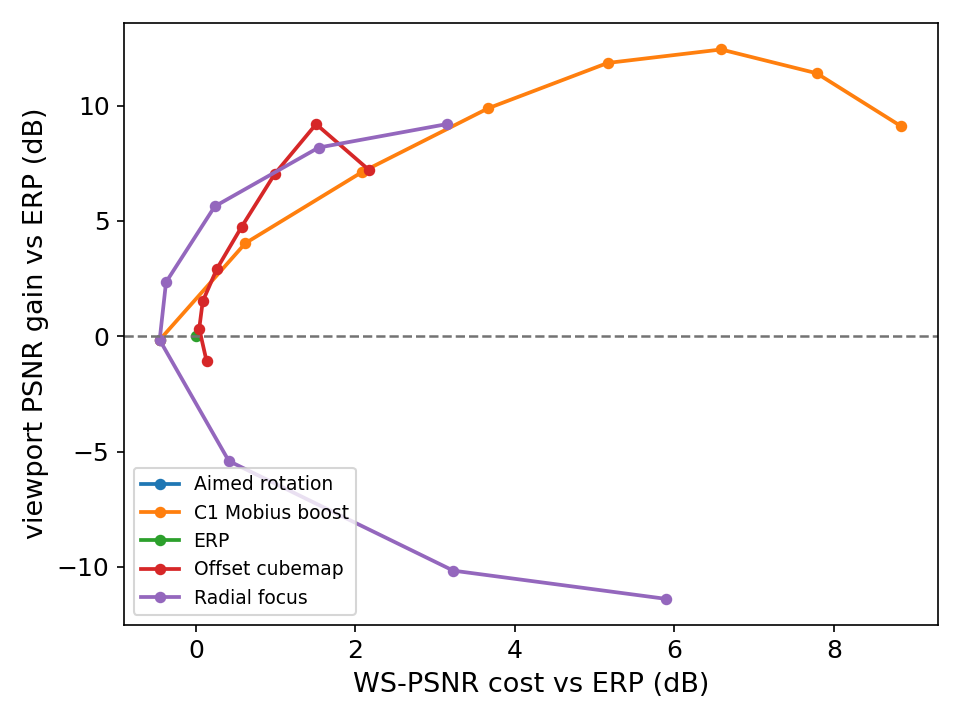}
\caption{Viewport/global-quality tradeoff.}
\end{subfigure}\hfill
\begin{subfigure}[t]{0.32\textwidth}
\centering
\includegraphics[width=\linewidth]{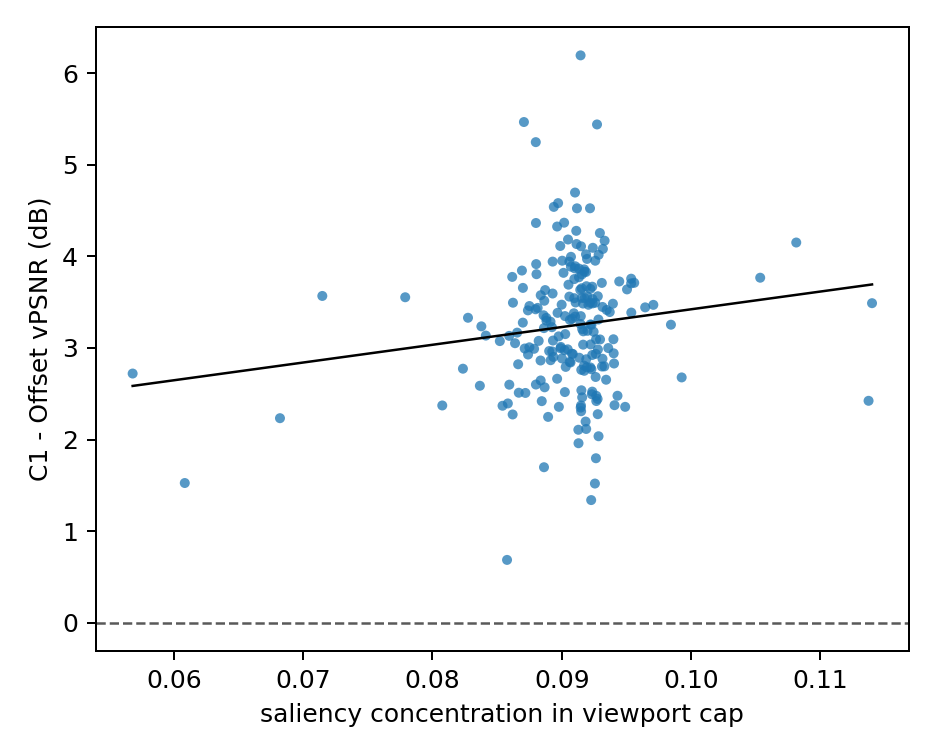}
\caption{Concentration and gain.}
\end{subfigure}
\caption{Expanded empirical picture. C1 wins the paired viewport-PSNR comparison against offset cubemap across all 216 cases, but it occupies a specific Pareto region: higher viewport quality at higher WS-PSNR cost. In the SUN360 saliency-targeted regime, gains remain consistently positive across the observed concentration range; the broader concentration effect is better illustrated by real-gaze window length in Table~\ref{tab:gaze}.}
\label{fig:expanded}
\end{figure*}

\subsection{Pareto tradeoff rather than a free win}
The expanded audit supports a more honest claim than ``C1 always wins.'' C1 is a strong viewport-quality and shear-free point, but it spends more of the global budget near the target. Table~\ref{tab:pareto} controls for this by matching search budget, center scale, and WS-PSNR cost.

\begin{table*}[t]
\centering
\caption{Pareto matching between C1 and optimized offset cubemap on 216 paired cases. Positive vPSNR delta means C1 is better in the target viewport.}
\label{tab:pareto}
\small
\setlength{\tabcolsep}{4pt}
\resizebox{\textwidth}{!}{%
\begin{tabular}{lcccc}
\toprule
Matching condition & Cases & C1 - Offset vPSNR & Median & Diagnostic \\
\midrule
Same scalar-search budget & 216 & $+3.24{\pm}0.05$ & $+3.26$ & 216/216 wins \\
Same center forward scale & 216 & $+0.70{\pm}0.06$ & $+0.70$ & scale gap 0.89 \\
Same WS-PSNR cost as optimized Offset & 216 & $-4.74{\pm}0.10$ & $-4.79$ & WS gap 0.75 dB \\
Same viewport-PSNR target as Offset & 216 & \na & \na & reachable 216/216; $+2.43{\pm}0.06$ dB more WS cost \\
\bottomrule
\end{tabular}}
\end{table*}

The first two rows show that C1 is better when the comparison is matched by scalar-search budget or local center scale. The third and fourth rows expose the cost: if full-sphere WS-PSNR is held fixed, optimized offset is better for the target viewport, and matching offset's viewport PSNR requires extra global sacrifice. A simple application utility makes this tradeoff explicit. Let
\begin{equation}
U_\alpha = \alpha\,\mathrm{vPSNR} + (1-\alpha)\,\mathrm{WSPSNR},
\end{equation}
where $\alpha$ is the system weight placed on the current viewport. This scalarization is not intended as a perceptual QoE model; it is a diagnostic summary of the measured viewport/global tradeoff. Using the default C1--Offset deltas in Table~\ref{tab:paired}, $\Delta \mathrm{vPSNR}=+3.24$ dB and $\Delta \mathrm{WSPSNR}=-4.88$ dB, C1 is preferred when
\begin{equation}
\alpha > \frac{4.88}{4.88+3.24}\approx 0.60.
\end{equation}
Thus C1 is appropriate when the application values the current viewport at roughly 60\% or more of the utility, while offset cubemap is preferable when global uniform quality dominates.

\begin{figure}[t]
\centering
\includegraphics[width=\linewidth]{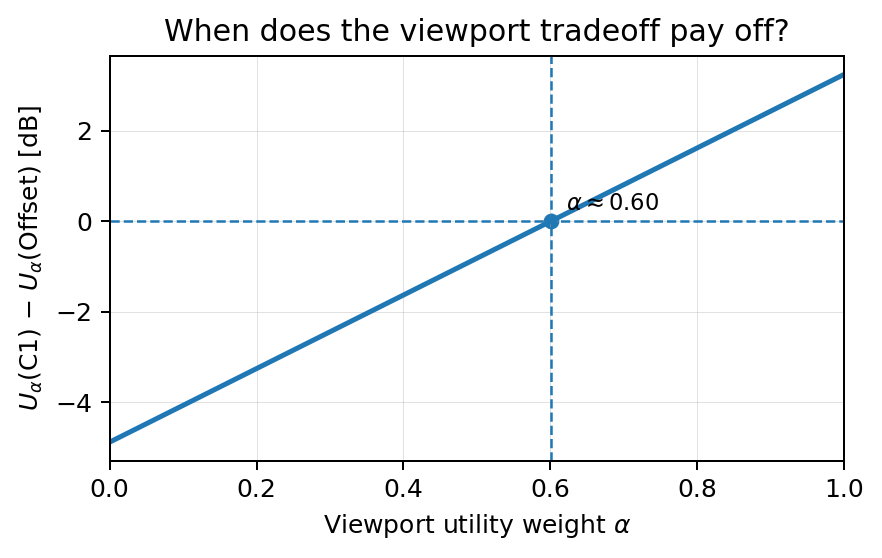}
\caption{Diagnostic utility view of the Pareto tradeoff. The break-even point occurs near $\alpha=0.60$ for the measured C1--Offset viewport and WS-PSNR deltas; this is not a perceptual QoE model.}
\label{fig:utility}
\end{figure}

\subsection{Geometric distortion and sampling behavior}
Table~\ref{tab:distortion} separates visual fidelity from geometric distortion. C1 has no angular/shear distortion, but it still changes scale and area. This distinction is central to the paper's claim.

\begin{table}[t]
\centering
\caption{Geometric behavior of the main comparison maps. $\Kqc$ measures local anisotropy/shear, not scale or area change; C1 is shear-free but not area-preserving.}
\label{tab:distortion}
\small
\setlength{\tabcolsep}{3.5pt}
\resizebox{\linewidth}{!}{%
\begin{tabular}{lccccc}
\toprule
Warp & Center scale & Median scale & $\Kqc_{p95}\downarrow$ & $\Kqc_{max}\downarrow$ & Notes \\
\midrule
Aimed rotation & 1.00 & 1.00 & 1.00 & 1.00 & isometry \\
Offset cubemap & 11.06 & 0.64 & 2.29 & 2.43 & non-conformal \\
\textbf{C1 M\"obius boost} & 15.75 & 0.12 & \textbf{1.00} & \textbf{1.00} & conformal \\
\bottomrule
\end{tabular}}
\end{table}

\subsection{Prediction uncertainty and gated deployment}
Table~\ref{tab:gated} evaluates the deployment failure mode directly. Strong C1 is effective when the target is accurate, but it becomes unsafe near $30\degree$ prediction error. The error-conditioned gated study quantifies which boost strength is appropriate as target uncertainty increases; it motivates, but does not replace, a real predictor-confidence mechanism.

\begin{table}[t]
\centering
\caption{Error-conditioned gating study on 72 SUN360 images. This is not a full viewport predictor; it quantifies which boost strength is appropriate as target uncertainty increases. Failure means the policy underperforms ERP on the true viewport.}
\label{tab:gated}
\small
\setlength{\tabcolsep}{3.2pt}
\resizebox{\linewidth}{!}{%
\begin{tabular}{ccccc}
\toprule
Error & Strong C1 gain & Strong fail & Gated C1 gain & Gated fail \\
\midrule
$0\degree$ & $+11.46{\pm}0.21$ & 0.00 & $+11.46{\pm}0.21$ & 0.00 \\
$5\degree$ & $+10.62{\pm}0.25$ & 0.00 & $+10.62{\pm}0.25$ & 0.00 \\
$10\degree$ & $+8.44{\pm}0.28$ & 0.00 & $+8.44{\pm}0.28$ & 0.00 \\
$20\degree$ & $+3.51{\pm}0.26$ & 0.04 & $+7.61{\pm}0.18$ & 0.00 \\
$30\degree$ & $-0.35{\pm}0.20$ & 0.64 & $+5.26{\pm}0.19$ & 0.00 \\
$45\degree$ & $-4.14{\pm}0.16$ & 1.00 & $+1.65{\pm}0.15$ & 0.08 \\
\bottomrule
\end{tabular}}
\end{table}

\subsection{Filtering and anti-aliasing}
Table~\ref{tab:filtering} verifies that the C1/offset gap is not a single interpolation artifact. The C1 advantage remains positive across interpolation and prefiltering choices, but it shrinks under aggressive low-pass filtering. This indicates that much of the benefit comes from preserving high-frequency viewport detail, rather than from a universal low-frequency reconstruction advantage.

\begin{table}[t]
\centering
\caption{Filtering and anti-aliasing check on 24 SUN360 images. Values are paired C1 - Offset viewport-PSNR gains.}
\label{tab:filtering}
\small
\setlength{\tabcolsep}{4pt}
\resizebox{\linewidth}{!}{%
\begin{tabular}{lcccc}
\toprule
Filter / reconstruction & Mean gain & Median & Win rate & $n$ \\
\midrule
Bilinear & $+2.42{\pm}0.14$ & $+2.43$ & 24/24 & 24 \\
Bicubic & $+2.69{\pm}0.23$ & $+2.67$ & 24/24 & 24 \\
Gaussian prefilter $\sigma=0.8$ & $+0.70{\pm}0.04$ & $+0.71$ & 24/24 & 24 \\
Gaussian prefilter $\sigma=1.2$ & $+0.31{\pm}0.02$ & $+0.30$ & 24/24 & 24 \\
\bottomrule
\end{tabular}}
\end{table}

\subsection{Attention regime and small real-gaze illustration}
Table~\ref{tab:gaze} records a small Salient360 trace study. It is an illustrative operating-regime check rather than a full gaze benchmark: short per-user windows are concentrated and produce large gains, while all-user aggregate saliency is diffuse and gives only a small gain.

\begin{table}[t]
\centering
\caption{Illustrative operating regime on one Salient360 panorama with three users. Short concentrated windows produce large gains; diffuse all-user aggregation does not.}
\label{tab:gaze}
\small
\setlength{\tabcolsep}{3.5pt}
\resizebox{\linewidth}{!}{%
\begin{tabular}{lccccc}
\toprule
Regime / window & Users & Windows & Conc. median & C1 gain median & $n$ \\
\midrule
0.5 s instantaneous & 3 & 15 & 1.00 & $+17.98$ & 15 \\
1.0 s instantaneous & 3 & 15 & 1.00 & $+16.62$ & 15 \\
2.0 s short-term & 3 & 15 & 0.73 & $+8.29$ & 15 \\
4.0 s short-term & 3 & 15 & 0.45 & $+3.90$ & 15 \\
All-user aggregate & 3 & 1 & 0.19 & $+0.99$ & 1 \\
\bottomrule
\end{tabular}}
\end{table}

\subsection{Implementation overhead}
The encoder-side scalar search is not the right deployment-time bottleneck because maps can be cached once the target and strength are known. The C1 protocol transmits four scalar parameters, matching offset cubemap in side-information size. Appendix~\ref{app:runtime} reports a CPU timing sanity check for cached-map decoding; because these timings are hardware- and implementation-dependent, they are not used as central evidence for the geometric claim.

\FloatBarrier
\section{Discussion}
The intended interpretation is geometric rather than universal. A M\"obius boost is exactly conformal, so it is shear-free, but it is not distortion-free in every sense: it changes local scale and area, and it necessarily sacrifices samples away from the attended region. The expanded results make this explicit. Under the same scalar-search budget, C1 improves viewport PSNR over offset cubemap on all 216 paired cases and all 72 image-level aggregates. Under the same WS-PSNR cost, however, it loses to offset cubemap. This is the correct framing: C1 is a Pareto point for applications that value the predicted viewport more than global uniform quality.

Prediction uncertainty is the main deployment risk. Strong magnification is useful only when the target is accurate. The error-conditioned gating study turns the large-error failure into an operating rule: use strong C1 for reliable targets, weaken the boost for moderate uncertainty, and fall back when uncertainty is high or attention is diffuse. A real deployment should implement this with predictor confidence or uncertainty rather than test-time ground-truth error. The small Salient360 trace study supports the same qualitative regime: short per-user windows are concentrated, whereas aggregate all-user saliency behaves like an encode-once signal and yields only a small benefit.

The non-conformal radial focus baseline and C2 are secondary stress tests, not the main evidence. Offset cubemap is the primary literature-grounded non-conformal opponent. The appendix shows that C2 helps in synthetic far-apart two-target cases, but it loses exact M\"obius structure and introduces measurable quasiconformal distortion. Unless a target application contains real far-apart multi-attention events, exact C1 should be the main primitive and any multi-center map should be selected only by an encoder-side objective with distortion reporting.

\section{Limitations and Future Work}
This paper isolates fixed-sample geometric resampling rather than a complete streaming system. We therefore do not claim codec-level bitrate savings, tile-manifest optimization, or trajectory-level QoE improvement. The main evaluation uses saliency-derived static viewport targets on SUN360; real fixation-level behavior is illustrated through a small Salient360 trace study rather than a full gaze benchmark. Offset cubemap is the primary literature-grounded non-conformal baseline, while the radial focus warp is included as an additional non-conformal audit baseline. Finally, downstream probes are lightweight sanity checks rather than evidence for broad task-level computer-vision benefits.

\section{Conclusion}
We frame viewport-adaptive 360\degree{} resampling as shear-free magnification on the sphere. Spherical M\"obius boosts provide an exact conformal map that can concentrate samples around a target viewport while preserving local angles. The expanded experiments show a consistent viewport-PSNR advantage over optimized offset cubemap under a fixed buffer and scalar-search budget, while also exposing the necessary global WS-PSNR tradeoff and the need for uncertainty-aware deployment. The resulting primitive is not a universal 360\degree{} layout, but a clean geometric building block for prediction-conditioned foveated 360\degree{} imaging.

\appendix
\section{Additional Experiments}
The following probes are included for completeness. They are not central to the main C1 claim and should be interpreted as supplementary evidence.

\subsection{Downstream local recovery probe}
Table~\ref{tab:downstream} reports two lightweight downstream probes on reconstructed attended viewports. These probes are lightweight sanity checks rather than evidence for broad task-level computer-vision benefits; they only check whether the geometric fidelity gains preserve recoverable local structure in this limited setting.

\begin{table}[H]
\centering
\caption{Downstream recovery on local/ImageNet viewport crops. All methods use the same fixed sample budget; $n=5$ crops.}
\label{tab:downstream}
\small
\setlength{\tabcolsep}{3.5pt}
\resizebox{\linewidth}{!}{%
\begin{tabular}{lcccc}
\toprule
Metric & ERP & Offset & Radial & \textbf{C1} \\
\midrule
SIFT inliers $\uparrow$ & 12.6 & 100.6 & 96.4 & \textbf{150.4} \\
Localization error px $\downarrow$ & 1.23 & 0.70 & 0.79 & \textbf{0.54} \\
Feature cosine $\uparrow$ & 0.615 & 0.864 & 0.877 & \textbf{0.910} \\
ImageNet top-1 agreement $\uparrow$ & 20\% & \textbf{80\%} & 40\% & 60\% \\
\bottomrule
\end{tabular}}
\end{table}

\subsection{Resolution and FoV sanity checks}\label{app:sensitivity}
Table~\ref{tab:sensitivity} reports a smaller sanity check on four high-resolution real-image cases ($H=512$ for the source image in these runs). These results are not intended to replace the 72-image SUN360 audit. The purpose is only to check that the C1--offset gap does not disappear under larger buffers or common viewport FoVs.

\begin{table}[H]
\centering
\caption{Resolution/FoV sanity checks. Values are C1 minus the stronger non-conformal baseline on the same case.}
\label{tab:sensitivity}
\small
\setlength{\tabcolsep}{3.5pt}
\resizebox{\linewidth}{!}{%
\begin{tabular}{llccc}
\toprule
Ablation & Setting & Mean gain & Median & $n$ \\
\midrule
Buffer resolution & $120\times60$ & $+2.96{\pm}0.35$ & $+2.94$ & 4 \\
Buffer resolution & $240\times120$ & $+3.13{\pm}0.36$ & $+3.06$ & 4 \\
Buffer resolution & $480\times240$ & $+3.13{\pm}0.32$ & $+3.02$ & 4 \\
Viewport FoV & $30\degree$ & $+3.75{\pm}0.22$ & $+3.79$ & 4 \\
Viewport FoV & $60\degree$ & $+2.84{\pm}0.28$ & $+2.74$ & 4 \\
Viewport FoV & $90\degree$ & $+2.09{\pm}0.26$ & $+2.04$ & 4 \\
Viewport FoV & $110\degree$ & $+1.48{\pm}0.22$ & $+1.45$ & 4 \\
\bottomrule
\end{tabular}}
\end{table}

\subsection{Cached-map runtime sanity check}\label{app:runtime}
Table~\ref{tab:runtime} reports CPU-only cached-map decoding timings with input $H=512$, buffer $240\times120$, ten repeats, and no file I/O. ``Cache build'' constructs the dense sampling map after the method parameter is known; ``cached decode'' applies the cached map. The absolute numbers depend on the unoptimized Python/NumPy implementation and hardware, so the table is used only as an overhead sanity check.

\begin{table}[H]
\centering
\caption{Cached-map runtime and side information in the CPU prototype. Timings exclude file I/O; cached decode excludes map construction.}
\label{tab:runtime}
\small
\setlength{\tabcolsep}{3.2pt}
\resizebox{\linewidth}{!}{%
\begin{tabular}{lccccc}
\toprule
Method & Params & Side bits & Store ms & Cache build ms & Cached decode ms \\
\midrule
Aimed rotation & 3 & 96 & 12.5 & 88.2 & 84.8 \\
Offset cubemap & 4 & 128 & 9.5 & 52.7 & 81.0 \\
Radial focus warp & 4 & 128 & 11.0 & 69.1 & 85.6 \\
\textbf{C1 M\"obius boost} & 4 & 128 & 9.3 & 89.4 & 84.8 \\
\bottomrule
\end{tabular}}
\end{table}

\subsection{Multi-center quasiconformal extension details}\label{app:c2}
A single exact boost has one magnification center. To study simultaneous far-apart attention clusters, we include a windowed multi-center map defined as the time-one flow of localized boost-like vector fields,
\begin{align}
V(d)&=\sum_{k=1}^{K} s_k\,w_k(d)\,\big(c_k-(c_k\!\cdot d)d\big),\label{eq:flow}\\
w_k(d)&=\exp\{\kappa(c_k\!\cdot d-1)\}.
\end{align}
The unwindowed one-center case ($K=1,\kappa=0$) integrates to an exact M\"obius boost. With a nonconstant window, however, even a one-center flow is generally not an exact M\"obius map. We therefore treat C1 and C2 as separate encoder candidates,
\begin{equation}
\mathcal{G}=\mathcal{G}_{\mathrm{C1}}\cup\mathcal{G}_{\mathrm{C2}},
\end{equation}
and select C2 only when it improves the measured objective under a distortion constraint. For this preliminary extension, the appendix reports $\Kqc_{p95}$ and selection gain as regime diagnostics; it does not make a deployment claim for multi-center mapping.

\subsection{C2 extension: multi-center attention}
C2 is framed as an extension rather than the main contribution. Table~\ref{tab:c2} reports the controlled synthetic separation sweep and shows the regime where a multi-center map becomes useful: attention components must be far apart.

\begin{table}[H]
\centering
\caption{C2 multi-center extension. Selection gain means encoder-side selection between exact C1 and windowed C2. C2 helps primarily when two attention components are separated by large angles.}
\label{tab:c2}
\small
\setlength{\tabcolsep}{4pt}
\resizebox{\linewidth}{!}{%
\begin{tabular}{lcccccc}
\toprule
Case & Separation & C1 gain & C2 gain & Selection gain & $\Kqc_{p95}$ C2 & $n$ \\
\midrule
Synthetic two-object & $0$--$30\degree$ & $13.16$ & $6.04$ & $0.00$ & $2.45$ & 1 \\
Synthetic two-object & $30$--$60\degree$ & $5.56$ & $4.80$ & $0.00$ & $2.57$ & 1 \\
Synthetic two-object & $60$--$90\degree$ & $1.83{\pm}0.92$ & $8.27{\pm}0.97$ & $6.44{\pm}1.89$ & $2.68{\pm}0.17$ & 2 \\
Synthetic two-object & $>90\degree$ & $0.64{\pm}0.32$ & $9.74{\pm}0.39$ & $9.11{\pm}0.23$ & $3.15{\pm}0.32$ & 3 \\
\bottomrule
\end{tabular}}
\end{table}

\end{document}